%% file: acl_latex.tex
\pdfoutput=1

\documentclass[10.5pt]{article}

\usepackage[]{acl}

\usepackage{times}
\usepackage{latexsym}

\usepackage[T1]{fontenc}

\usepackage[utf8]{inputenc}

\usepackage{microtype}
\usepackage{graphicx}
\usepackage{float}

\title{Linguistic Knowledge in Data Augmentation for Natural Language Processing: An Example on Chinese Question Matching}

\author{Zhengxiang Wang \\
  Department of Linguistics, Stony Brook University \\
  \texttt{zhengxiang.wang@stonybrook.edu}}

\begin{document}
\maketitle
\thispagestyle{plain}
\pagestyle{plain}
\begin{abstract}
To investigate the role of linguistic knowledge in data augmentation (DA) for Natural Language Processing (NLP), we designed two adapted DA programs and applied them to LCQMC (a Large-scale Chinese Question Matching Corpus) for a binary Chinese question matching classification task. The two DA programs produce augmented texts by five simple text editing operations (or DA techniques), largely irrespective of language generation rules, but one is enhanced with a pre-trained n-gram language model to fuse it with prior linguistic knowledge. We then trained four neural network models (BOW, CNN, LSTM, and GRU) and a pre-trained model (ERNIE-Gram) on the LCQMC’s train sets of varying size as well as the related augmented train sets produced by the two DA programs. The results show that there are no significant performance differences between the models trained on the two types of augmented train sets, both when the five DA techniques are applied together or separately. Moreover, due to the inability of the five DA techniques to make strictly paraphrastic augmented texts, the results indicate the need of sufficient amounts of training examples for the classification models trained on them to mediate the negative impact of false matching augmented text pairs and improve performance, a limitation of random text editing perturbations used as a DA approach. Similar results were also obtained for English. 
\end{abstract}

\section{Introduction}

Data augmentation (DA) is a common solution to the problems of limited and imbalanced data. It works by generating novel and label-preserving data from the existing data \cite{xie2020unsupervised}, which would otherwise be unavailable or expensive to collect. Owing to the increasing popularity of supervised deep learning models that demand large-scale labeled data as well as more studies on understudied/under-resourced language and text domains, the Natural Language Processing (NLP) community has seen a growing interest in DA in recent years \cite{feng-etal-2021-survey,Liu:9240734,Shorten2021}. However, unlike image and speech, whose physical features can be relatively easily manipulated without deviating from the original labels, text augmentation poses a bigger challenge. This is simply because there is no easy and automatic way to paraphrase a randomly given piece of text while preserving its linguistic integrity and, above all, meaning. As such, while there are well established and widely applied DA techniques as well as frameworks in image and speech recognition research\footnote{ Although there is certain overlap between speech recognition and NLP, they are two independent fields with divergent concerns and specializations \cite{manning99foundations}. Typically, NLP is about text processing only.}  with noteworthy success \cite{Iwana2021,Park2019,Shorten2019ASO}, DA for NLP as a whole remains underexplored \cite{feng-etal-2021-survey}.

The main purpose of this paper is to investigate a fundamental question we found unanswered to the best of our knowledge: the role of linguistic knowledge in DA for NLP; in particular, whether more linguistic knowledge leads to a better DA approach. By a better DA approach, we mean one that can lead to superior trained models’ performance on a given NLP task. Intuitively, with more linguistic knowledge instilled, a DA approach is expected to augment text of higher-quality or more grammatical and thus to be presumably better. We believe a deeper understanding of what counts as a better DA approach and the role of linguistic knowledge will trigger more in-depth experiments and discussions and advance this research area to the next stage. Eventually, these efforts will turn into potential great benefits, both academically and commercially, helping train robust NLP models with small data. 

To conduct our research, we present two DA programs and train five supervised classification models on the augmented train sets for a binary Chinese question matching classification task. For simplicity and interpretability concerns, the DA programs used in this study are adapted from the Easy Data Augmentation (EDA) program \cite{wei-zou-2019-eda}, which augments text by four naïve text editing operations, largely irrespective of language generation rules. The only difference between the two adapted programs is whether they have a pre-trained statistical n-gram language model (LM) to select the most linguistically likely outputs, an effective mechanism to fuse a program with probabilistic linguistic knowledge. We choose n-gram LM over neural LMs because it is more efficient to train, and most importantly, more interpretable for its straightforward frequency-based approach. As the EDA approach has shown success \cite{wei-zou-2019-eda} in various sentiment-related and sentence type classification tasks with small datasets (e.g., mostly around 10k examples), we choose LCQMC (a Large-scale Chinese Question Matching Corpus) compiled by \citet{liu-etal-2018-lcqmc} to compare the goodness of the two adapted programs, a large labeled corpus with over 260k examples. Since our corpus is much larger and the question matching task involves comparing a pair of text, instead of one, for label prediction, it is a more reliable way to test the capacity and generalizability of a DA approach. In principle, if a DA approach can work well for the question matching task, it should also show promise for those simpler and related NLP tasks, as question matching, or text matching, is one of the most basic tasks for NLP. 

The contributions of this paper are threefold. First, we present the first study on the role of linguistic knowledge in DA for NLP with a special focus on the effects of probabilistic linguistic knowledge on a DA approach or technique. Second, we propose two DA programs adapted from the EDA program. Although the adapted programs are for augmenting Chinese, several changes we made, including a new DA technique and the added n-gram LM, can be universal for tailoring the EDA program to other languages. Third, we also fill the research gaps in two understudied areas: DA for question matching classification task and DA for Chinese NLP. 

The code, data, and results for this study are available at \url{https://github.com/jaaack-wang/linguistic-knowledge-in-DA-for-NLP}.

\section{Related Works}

Thus far, various DA techniques has been employed in NLP research, such as thesaurus-based \cite{Zhang2015} and embedding-based \cite{wang-yang-2015-thats} word replacement, random text-editing perturbation \cite{wei-zou-2019-eda}, rule-specific generation \cite{asai-hajishirzi-2020-logic, kang-etal-2018-adventure}, back translation \cite{sennrich-etal-2016-improving, Singh2019}, and neural-model-based predictive text transformation \cite{hou-etal-2018-sequence, kobayashi-2018-contextual, kurata16b_interspeech} etc. Most of these studies find slight but stable performance gains for training models with augmented data for given NLP tasks, such as text classification, question answering, machine translation, for a common reason that the augmented data introduces noise to the original train set and prevents the trained models from overfitting, which improves the models’ generalizability on the test set. 

As the NLP community is more engaged in exploring the usefulness of DA for specific NLP tasks, we have not been able to find any focused studies from the existing literature related to the subject matter of this study, i.e., the role of linguistic knowledge in DA for NLP. However, some indirect evidence seems to be affirmative. For example, \citet{kobayashi-2018-contextual} trained a recurrent neural network (RNN) LM, which replaces words with paradigmatic relations predicted by the RNN LM to generate new examples. Since this approach ignores the semantic association between the replaced words and the corresponding labels, he also constrained the LM to predict words more compatible with the given labels by probability. By so doing, he found about 0.2\% overall improvements in accuracy for 5 sentiment-related and one question type classification tasks. According to the results reported by \citet{kang-etal-2018-adventure}, we also find that while not consistently, a sequence to sequence (seq2seq) DA model blended with a few hand-crafted rules increases more test set accuracy than the base seq2seq DA model when certain ratios of two textual inference datasets were augmented. However, since these neural DA models already encode and learn implicit linguistic knowledge through complex representation learning, it is not possible to fully recognize the effects of those added linguistic knowledge, either implicit or explicit, in them.  

Relevant to our hypothesis on what counts as a better DA approach, we can find strong supports by thinking in reverse. That is, although text augmentation helps increase the size of the training texts, which then improves the performance of the trained models through regularization, it is still incomparable to the human-produced-and-annotated training texts of a same size, which by default we assume to be superior in quality as well as more diverse. For example, in  \citet{wei-zou-2019-eda}, they augmented the original training examples by a factor of 9, giving them 5,000 training examples when 500 were given. Although the augmented train set shows average 3\% performance gains in accuracy on the test set for 5 classification tasks, compared to that without augmentation, this is still significantly lower than the average 10\% performance improvements when the models are trained on 5,000 of the original training examples\footnote{ \citet{wei-zou-2019-eda} claims that with the augmented texts, their classification models achieve higher average accuracy using only 50\% of the train set than when the models are trained on the entire train set without augmentation. This is misleading since the performance of their models starts plateauing when the models see 20\% of the train set.}. Therefore, we expect that coupled with a n-gram LM, the adapted EDA program that utilizes random text-editing perturbations, will augment higher-quality text, and thus achieve better trained models’ performance.

\section{Experimental Setup}

\subsection{LCQMC}

LCQMC contains over 260k question pairs, extracted from BaiduKnows, a Quora-like online Q\&A platform. Each question pair is manually annotated by three external professional annotators with a label, 1 or 0, to represent whether a question pair matches or not in terms of the expressed intents. As judgements vary from person to person and the interpretation of some question pairs is bound to contexts, there are about 15\% annotation inconsistency and 20\% annotation uncertainty \cite{liu-etal-2018-lcqmc}. In this study, we keep the original separation of the train set, the development set, and the test set as is in LCQMC, whose basic statistics are shown in Table~\ref{tab:data}.

\input{tables/lcqmc_data.tex}

\subsection{Two adapted DA programs}

The base DA program developed in this study is adapted from the EDA program\footnote{\url{https://github.com/jasonwei20/eda_nlp/tree/master/code}.} \cite{wei-zou-2019-eda} and the control DA program is the base program combined with a pre-trained statistical n-gram LM (refer to the next section). We name these two programs as the REDA program and the $REDA_{+NG}$ program respectively, where REDA stands for Revised Easy Data Augmentation.

Like the EDA program, the REDA program also has four text editing operations, i.e., Synonym Replacement (SR), Random Swap (RS), Random Insertion (RI), and Random Deletion (RD). Their functions are as follows: SR works by randomly replacing synonyms for eligible words based on a given dictionary, while RS works by randomly swapping word pairs. RI inserts random synonyms, if any, instead of random words, to avoid uncontrolled label change. In contrast, RD deletes words at random. We used jieba\footnote{\url{https://github.com/fxsjy/jieba}.}, a popular Chinese text segmentation tool, to tokenize Chinese text throughout this research. 

To further diversify the augmented texts, we also created a new text editing operation called Random Mix (RM), which randomly selects 2-4 of the other four operations to produce novel texts. Besides, a few major changes were also made to fix few bugs we found on the EDA program and to better serve our needs of augmenting Chinese and conducting this research, including:

\begin{enumerate}
    \itemsep0em
    \item We rewrote the entire program to ensure that there are no duplicates in the augmented texts, including one for the original text. Duplicates can occur when there are no synonyms to replace (SR) or insert (RS) for words in the original texts, or when the same words are replaced or swapped back during SR and RS operations. 
    \item The REDA program does not preprocess the input text by removing punctuations or by introducing stop words. We did not find this type of preprocessing helpful and necessary in general or makes sense for the basic idea of random text editing behind the EDA program. 
    \item Instead of using WordNet for SR, we compiled a preprocessed Chinese synonym dictionary leveraging multiple reputational sources\footnote{\url{https://github.com/jaaack-wang/Chinese-Synonyms}.}, including Chinese Open Wordnet\footnote{\url{http://compling.hss.ntu.edu.sg/cow/}.}. Moreover, unlike the EDA program, the REDA program only replaces one word at a given position at a time, instead of replacing all its occurrences, which we see as extra edits. 
\end{enumerate}

The $REDA_{+NG}$ program inherits the base REDA program but additionally utilizes the n-gram LM pre-trained to select the most likely augmented text(s) for each text editing operation from a variety of possible outputs. We have open-sourced two separate versions of code for these two DA programs, but during this study, we always combined them together in one working procedure so that the augmented texts outputted by these two programs are selected from the same pool. The implementation of this combination is also available at the open-sourced GitHub repository.

\subsection{N-gram LM}

To train the n-gram LM, we first compiled an independent corpus of BaiduKnows Q\&A texts based on an existing project found on GitHub, which scrapes over 9 million question-answer pairs from BaiduKnows platform\footnote{\url{https://github.com/liuhuanyong/MiningZhiDaoQACorpus}.}. This compiled corpus contains over 654 million words (or over 1.1 billion Chinese characters). Then, the relative frequency of unigram, bigram, trigram, and 4-gram for this corpus was calculated based on words and line by line with the results saved in four separate json dictionaries as the pre-trained parameters. When counting these n-grams, we added two special tokens, <START> and <END>, in the beginning and end of each line, to keep track of their tendency to stay ahead or at the end of a line. For efficiency concerns, we adjusted the relative frequency for the unigrams simply by assigning unseen vocabulary the same frequency with those one-off unigrams and employed stupid backoff without discounting unseen non-unigrams \cite{brants-etal-2007-large}. Finally, the n-gram LM takes the relative frequency of the n-grams as an estimation to their true probability of occurrence and calculates the maximum log probability of input text based on the chain rule of probability \cite{Jurafsky2009} as follows:

{\small $$\log P(NG_1: NG_n) = \log \prod_{i=1}^{n} P(NG_i) = \sum_{i=1}^{n} \log P(NG_i)$$}

\noindent 
where NG represents n-gram that is automatically generated by our n-gram LM. The n-gram starts with 4-gram, if any, and keeps backing off into low-order n-gram combination, if a higher-order n-gram is not available in the pre-made json dictionaries.

\subsection{Classification models}

We chose four neural network (NN) models and one transformer-based pre-trained model as the classification models. The NN models include the Bag of Words (BOW) model, the Convolutional Neural Network (CNN) model, and two RNN models: Long Short-Term Memory (LSTM) and Gated Recurrent Units (GRU). BOW model is a conventional technique to represent a text by summing up the embeddings of its words, and the similarity between texts is then often measured by Euclidean distance or cosine distance of the texts’ embeddings. Since \citet{kim-2014-convolutional}, CNN has been proven to be effective in various text classification tasks, including text pairing \cite{Severyn2015}. LSTM and GRU are two popular sequence models that consider word orders and have also been applied to semantic similarity tasks \cite{tai-etal-2015-improved, TIEN2019102090}, which we think may be especially useful for distinguishing the augmented texts from the natural texts, and more importantly, distinguishing the casually augmented texts by the REDA program from the conditionally augmented texts by the $REDA_{+NG}$ program in terms of the test set performance. Finally, the pre-trained model ERNIE-Gram \cite{xiao2020ernie-gram} was also chosen for its state-of-the-art performance on the LCQMC dataset.

The models were constructed using Baidu’s deep learning framework Paddle\footnote{\url{https://github.com/PaddlePaddle/Paddle}} and its NLP software PaddleNLP\footnote{\url{https://github.com/PaddlePaddle/PaddleNLP}}.

\section{Results}

\subsection{Quality of the augmented texts}

To evaluate the quality of the augmented texts generated by the REDA and $REDA_{+NG}$ programs, we designed three simple experiments to check their ability to restore to natural texts when modified texts or a pseudo synonym dictionary were given for three basic text editing operations, i.e., SR, RS, and RD. We skipped RI and RM because inserting random synonyms is generally not the natural way of language use however (un)natural the input text is and the text quality resulting from RM can be inferred from the other basic operations directly. 

The experiments went as follows. For SR, we designed a pseudo synonym dictionary made up of 3855 one-word-four-synonym pairs, where every word is mapped to four pseudo synonyms, one being the word itself and the rest non-synonym random words. All the words in the dictionary are those whose frequencies rank between the 1000th and the 10000th place in the unigram dictionary complied for the n-gram LM. For RS and RD, we randomly reordered the natural texts and added random words sampled from the texts respectively before RS and RD were performed. 10,000 pieces of texts were randomly sampled from the LCQMC’s train set for 5 times for every comparison we made. The average accuracy scores are reported in Table~\ref{tab:quality_experiment}.

\input{tables/quality_experiment.tex}

 As can be seen, while both programs’ performance declines as the number of edits increase, the $REDA_{+NG}$ program always outperform the REDA program in restoring to the natural texts. In fact, for the REDA program, restoring the modified texts to the original ones is a matter of chance equal to the inverse of the number of possible outputs available. However, the $REDA_{+NG}$ program augments texts of maximum likelihood, which tends to be closer to the natural texts expected. This is also true when natural texts are given as inputs. For example, through manual inspections, we found the $REDA_{+NG}$ program does much better in selecting the appropriate synonyms according to the linguistic contexts, which is a problem for the REDA program due to the ubiquitous existence of polysemy. By measuring the bigram overlap rate and edit distances of output texts randomly swapped twice from the natural texts, we found that the average overlap rate for the REDA program is much lower (i.e., 0.29 versus 0.77) and the average edit distances are much larger (i.e., 3.0 versus 1.4) than the $REDA_{+NG}$ program, meaning the latter preserves more collocational features of the natural texts and thus augments higher-quality texts.

Nevertheless, the $REDA_{+NG}$ program is also not free of considerable text quality decrease when more text edits are performed. This is largely due to the drastic increase of possible output texts as well as the more likely semantic shift of the original texts with large proportion of the input texts changed. Therefore, to conduct our research, the number of text edits performed is set proportional to the number of words of the input texts, so that a large quality difference of the augmented texts by the two programs can be maintained. More concretely, in the study, we set the SR and SR rate at 0.2 and the RI and the RD rate at 0.1 and applied Python rounding rules\footnote{When an even number ends with “.5”, it will be rounded down; otherwise, rounded up.}. RM will only randomly select two of the other four text editing operations with one text edit each for every input text to make the study more controlled. 

\input{tables/aug_data.tex}
\input{tables/accu.tex}

\subsection{Effects of the two DA programs }
\label{sec:da_experiment1}

We trained the five classification models in Baidu Machine Learning (BML) CodeLab on its AI Studio\footnote{\url{https://aistudio.baidu.com/aistudio/index}} with Tesla V100 GPU and 32GB RAM. The models were trained with 64 mini batches, a fixed 5e-4 learning rate (5e-5 for ERNIE-Gram model), and constantly 3 epochs. We used Adaptive Moment Estimation (Adam) optimizer and cross entropy loss function. We kept the original development set for validation purposes.

The following training sizes were experimented: 5k, 10k, 50k, 100k, and full size, approximately equal to 2\%, 4\%, 21\%, 42\%, and 100\% of the LCQMC’s train set respectively. When the train set size is 5k and 10k, we augmented two new texts for SR and RS, and one new text for RI, RD, and RM, because the last three text editing operations show smaller differences for the REDA and $REDA_{+NG}$ programs in terms of text quality (refer to the last section), which we want to hold as large as possible for the sake of this research. That translates into maximum 7 new texts for every text and up to 14 new texts for every text pair due to deduplication. Every augmented text was crossed paired with the other text that was a pair to the text being augmented with the original label kept for the newly made text pair. To make the training more manageable, we only augmented 5 new texts for every text with one output for every text editing operation, meaning a maximum tenfold increase in size when the associated train set size is 50k and more. The corresponding augmented train set size is given in Table~\ref{tab:aug_data}. 
The accuracy scores as well as the average precision, recall, and F1 scores on the test set are presented in Table~\ref{tab:accu} and Table~\ref{tab:metrics}, respectively. Contrary to our expectation, we do not find that the $REDA_{+NG}$ augmented train sets lead to better test set performance than the REDA augmented train sets, when it comes to the four metrics used in this study. According to the pairwise Mann-Whitney U tests we ran, there is no statistically significant difference across the four metrics among each type of models trained on the two types of augmented train sets, as the p-values were constantly far greater than .05. Although the former program does produce higher-quality augmented texts from a linguistic perspective as discussed above, evidence shows that models trained on the REDA augmented train sets outperform those trained on the $REDA_{+NG}$ augmented train sets by an average 0.3\% both in the accuracy and F1 scores. As can be seen from Table~\ref{tab:accu}, the $REDA_{+NG}$-led models only outperform the REDA-led ones in terms of the test set accuracy when the train set size is 5k for four models except the LSTM model and when the ERNIE-Gram models were finetuned on the full augmented train sets. Moreover, for any classification model trained on the REDA augmented train sets, in most cases, it achieves a slightly better score for the four metrics than the model trained on the $REDA_{+NG}$ augmented counterparts. It follows that the role of probabilistic linguistic knowledge instilled in the $REDA_{+NG}$ program is overall minimal and sometimes harmful to DA applied to the binary question matching task. 

\input{tables/metrics.tex}

Also noticeable from Table~\ref{tab:accu} is that 50k training examples appear to be the threshold where the two DA programs start bringing gains to the related test set accuracy scores compared to the baselines, except for the finetuned ERNIE-Gram models. However, as shown in Table~\ref{tab:metrics}, there is also a gap in the recall scores in favor of the baseline models, which may be attributed to the false matching text pairs produced by the two DA programs due to the inability of the underlying text editing operations to make strictly paraphrastic augmented texts. But these noisy augmented texts in return enable the classification models to generalize better on those matching text pairs judged to be non-matching by the baseline models, as indicated by the average larger precision scores. In addition, the advantage of the pre-trained model over the traditional NN models is significant: the ERNIE-Gram models, finetuned on all the three types of train sets, show about 12\% to 17\% average gains across the four metrics in relation to the other four trained models. This shows the promise of applying transfer learning to DA for NLP, which may be worth further studying in the future.

\subsection{Ablation study: each DA technique }

To gain a more nuanced understanding of the role of linguistic knowledge in each one of the DA techniques performed by the two DA programs, we conducted an ablation study where we trained models on train sets augmented by only one DA technique. That means, for a train set of given size randomly sampled from the LCQMC’s train set, there are five types of corresponding augmented train sets. Our analyses are based on comparing the average test set performance of the five models trained on the three types of train sets for the five augmentation scenarios. We also excluded ERNIE-Gram models, which are revealed to be distinct from the rest models across the four metrics in the last section, to see if there is a noticeable difference. 

As the training sizes are shown to have an effect on whether the DA-led models outperform the baseline models, to further validate that, we chose 11 training sizes for this ablation study, namely, 5k, 10k, 25k, 50k, 75k, 100k, 125k, 150k, 175k, 200k, and full set, roughly equal to 2\%, 4\%, 10\%, 21\%, 31\%, 42\%, 52\%, 63\%, 73\%, 84\%, and 100\% of the LCQMC’s train set respectively. The basic hyperparameters are same with the previous section. However, to make the training more manageable, we only trained 2 epochs when the baseline training size is 50~100k (included) and 1 epoch when the baseline training size is over 100k for the three types of train sets. Since it is evident from Table 4 that a larger training size under the same condition always leads to a higher test set performance, spending extra time in training a total of 605 models\footnote{Since there are 11 training sizes and 5 classification models, that translates into 55 models for the baseline train sets. As there are 5 DA techniques applied in 2 different ways (with or without n-gram LM), that translates into 550 (55 * 5 * 2) models for the augmented train sets. Hence, we have 605 models to train in total.}  with fixed 3 epochs may thus not be worthwhile to re-verify. Moreover, we only augmented 2 texts per text per DA technique when the baseline training size is no less than 50k and 1 text when otherwise, with the cross pairing applied, similar to what we did in the previous section. 
 
\begin{figure}[!htb]
\centering
\captionsetup{font=small}
  \includegraphics[width=1\columnwidth]{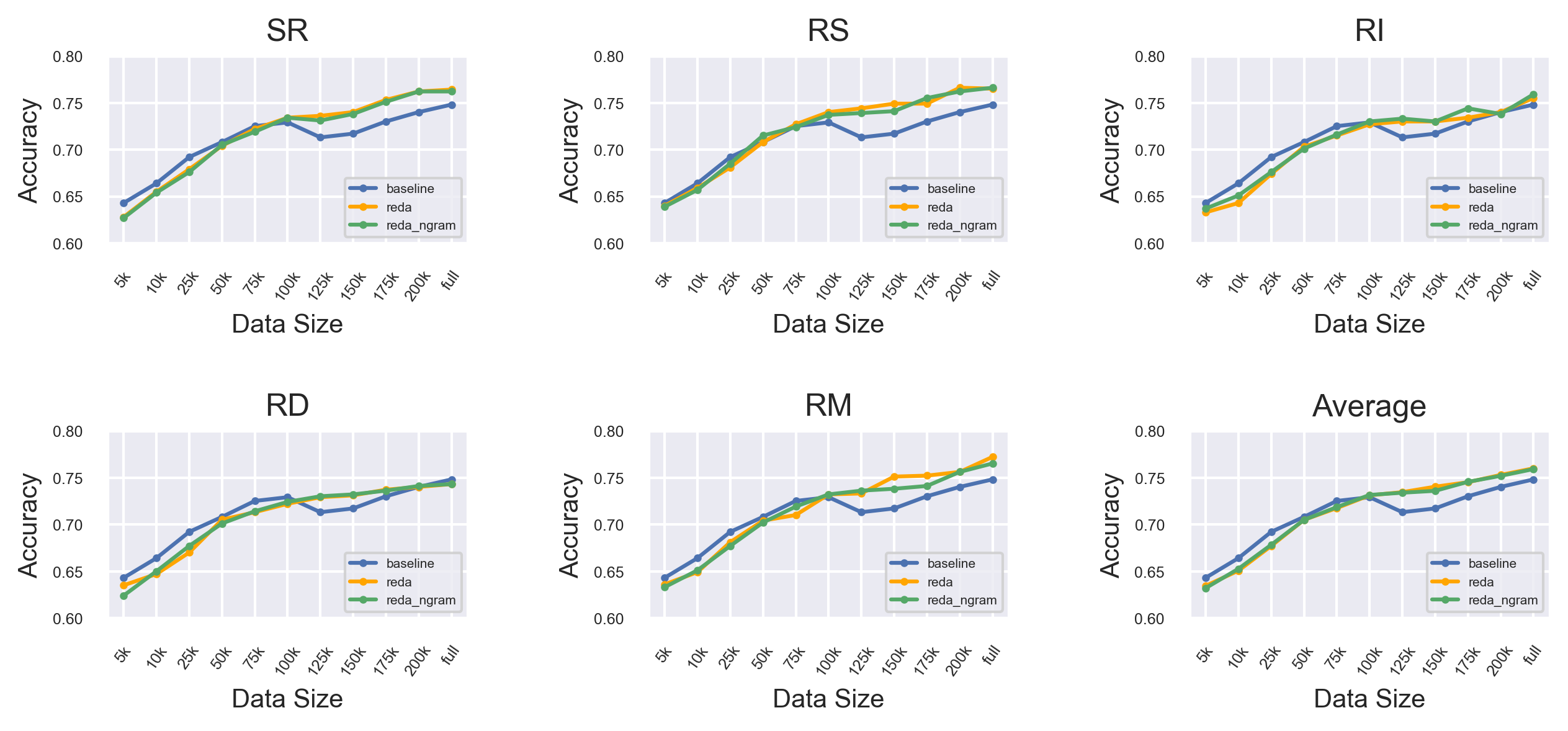}
\caption{Average test set accuracy scores of the three models under different conditions (i.e., text editing type, training data size) for the two types of LCQMC’s train sets. The sixth plot averages the statistics of the previous five plots.}

\label{fig:acc}
\end{figure}

Figure~\ref{fig:acc} shows the average test set accuracy scores of the five classification models trained on the three types of train sets under different text editing conditions and across different training sizes. In line with the previous finding, the effect of probabilistic linguistic knowledge on each one of the five DA techniques is minimal and of no statistically significant difference, both individually and on average. Although with certain text editing operations, such as RS, RI, and RM, there exist several points in which there is a relatively large difference in the accuracy scores between the two DA-led models, these differences fluctuate along the x-axis and eventually get reduced to be negligible when the average performance are concerned. This basic pattern remains true when we plotted the average test set performance based on any one of the four metrics with or without the ERNIE-Gram models\footnote{Details can be found in the GitHub project repository.}. 

\begin{figure}[!htb]
\centering
\captionsetup{font=small}
  \includegraphics[width=1\columnwidth]{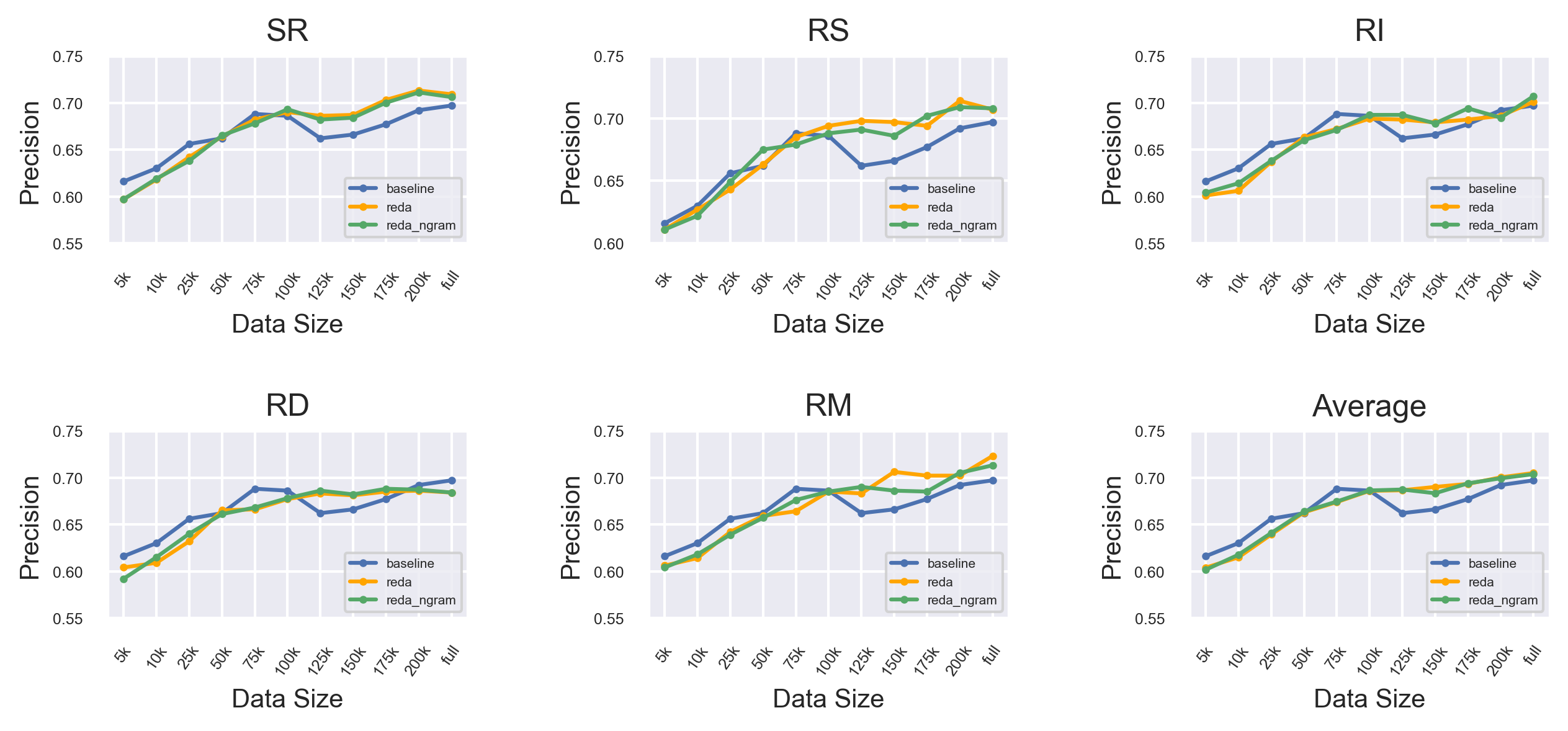}
\caption{Average test set precision scores of the five classification models.}

\label{fig:precision}
\end{figure}

\begin{figure}[!htb]
\centering
\captionsetup{font=small}
  \includegraphics[width=1\columnwidth]{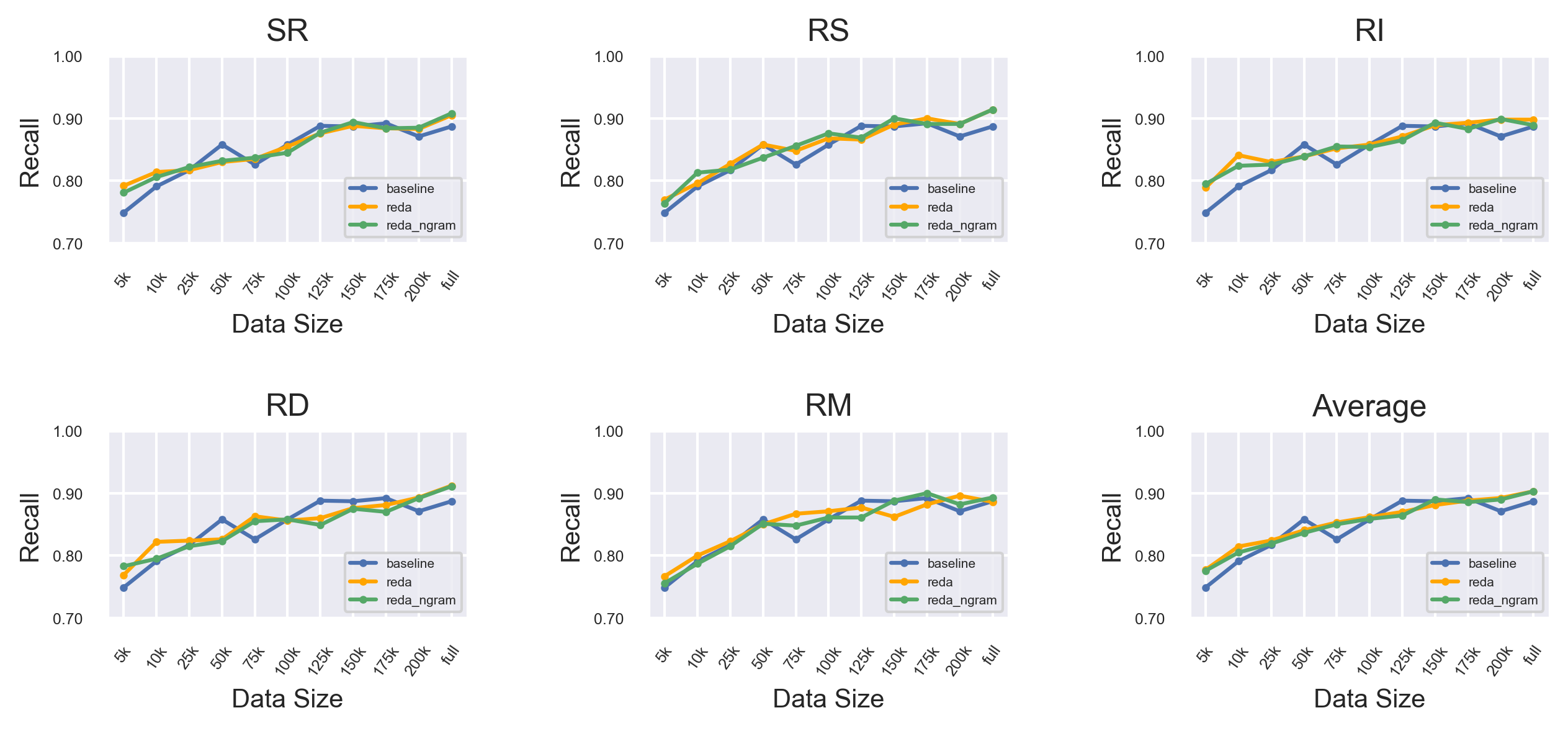}
\caption{Average test set recall scores of the five classification models.}

\label{fig:recall}
\end{figure}

Also related to the previous finding is that there does exist a threshold where the DA-led models outperform the baseline models in the test set accuracy scores, which appears to be the 100k training size or so, instead of 50k as in Table 4. The discrepancy may be explained by the different epoch numbers (e.g., 2 vs 3 for 50k) and possibly more importantly the separation of the DA techniques, which, however, are beyond the scope of this study. We also examined plots based on the other three metrics with or without the ERNIE-Gram models to explore the cause of such phenomenon. Figures~\ref{fig:precision} and~\ref{fig:recall} present the average test set precision and recall scores of the five classification models trained on the three types of train sets respectively. As can be seen, there is no general trend in which the baseline models surpass the DA-led counterparts in the test set recall scores, but a similar pattern that resembles that of Figure~\ref{fig:acc} also exists in Figure~\ref{fig:precision}. That means, the increase in the precision scores, after certain amounts of training examples are trained, are the main driver that makes the baseline models outperformed by the DA-led ones in terms of test set accuracy scores as well as the F1 scores, which are not shown here to save space. Moreover, this conclusion also largely holds when the ERNIE-Gram models are excluded.

\section{Discussions and Conclusions}

In this study, we examined the effects of linguistic knowledge on DA for a binary Chinese question matching task. We proposed two DA programs, i.e., the REDA and $REDA_{+NG}$ programs, that augment text by five random text editing operations (or DA techniques), with the $REDA_{+NG}$ program combined with a n-gram LM to fuse it with probabilistic linguistic knowledge. Surprisingly, we found that the $REDA_{+NG}$-led classification models did not surpass the REDA-led counterparts in the test set performance (i.e., accuracy, precision, recall, and F1 scores), which is also true when the five DA techniques in the two programs are applied and compared separately. In other words, our study indicates strongly that instilling more linguistic knowledge into a DA approach or technique does not necessarily make it a better one when it comes to training a better question matching classifier for Chinese, although doing so may make the augmented texts higher quality from a pure linguistic point of view. 

However, since the two DA-led models achieve very close scores in the four metrics with trivial advantages for the REDA-led models, it is not possible for us to explain why adding probabilistic linguistic knowledge as a constrain does not make a meaningful difference, positive or negative. A possible explanation might be that as the five deep learning models compare a pair of texts in vector space and the way how word embeddings encode linguistic knowledge is different from humans, performing simple text editing operations in two different ways (i.e., random, conditional) on a text may result in different meanings for humans, but that for machines nevertheless is less distinguishable in the high dimension of vector space. Moreover, as we only used probabilistic linguistic knowledge as a filter to select augmented texts closer to human language use, the inherent inability of the underlying text editing operations made by the two DA programs to produce strictly paraphrastic augmented texts means the two types of augmented texts are to a considerable extent comparable in that they are mostly not the paraphrases to the original texts being augmented. However, such interpretation cannot explain why the REDA-led models often outperform the $REDA_{+NG}$-led ones by a slight but consistent margin. 
 
\input{tables/qqqd_accu.tex}
\input{tables/qqqd_metrics.tex}

Unlike \citet{wei-zou-2019-eda} who show general success of their EDA program in bring performance gains for several sentiment-related and text type classification tasks across train sets of varying sizes, we only found such gains when the classification models were trained with sufficient amounts of training examples. As we expected in the beginning, question matching presents a more difficult and fundamental classification task because it involves comparing a pair of texts, instead of a single text, to predict the label for the given text pair. This nature makes question matching, or text matching in general, inherently much more sensitive to and subject to some tiny semantic changes caused by text augmentation. To further validate this hypothesis, we adjusted the two REDA programs and ran a post hoc experiment similar to Section~\ref{sec:da_experiment1} for English using the Quora Question Pairs Dataset (QQQD)\footnote{\url{https://quoradata.quora.com/First-Quora-Dataset-Release-Question-Pairs}}, from which we created three label-balanced data sets of comparable sizes to the LCQMC counterparts\footnote{Details can be found in the GitHub repository.}. The average test set accuracy scores in Table~\ref{tab:qqqd_accu} clearly show that models trained on the augmented train sets also need to see ample original training examples (near 150k or above) to stably outperform the baseline models, although the threshold is higher here. Therefore, for random text editing DA approach to work for question matching, there is a need of sufficient training examples to enable the trained models to mediate the negative impact of the false matching augmented text pairs resulting from random text editing perturbations and turn it into a means of regularization that improves the models’ generalizability. This is a general limitation of random text editing perturbations applied as a DA approach. 

Lastly, comparing the results from these two experiments, or between Table~\ref{tab:accu} and Table~\ref{tab:qqqd_accu}, and between Table~\ref{tab:metrics}  and Table~\ref{tab:qqqd_metrics}, we can see that the discussions and conclusions drawn from the LCQMC experiment mostly apply for the QQQD experiment as well, since the obtained data shares similar patterns. Besides the threshold difference noted above, which may be dataset specific, a noteworthy difference is that $REDA_{+NG}$-led models slightly but consistently outperformed the REDA-led counterparts of test set accuracy and precision, although there is also no statistically significant difference and the average F1 scores are same. This fact again demonstrates the difficulty of fully accounting for modern deep learning experiments, but it also strongly confirms the negligible role of probabilistic linguistic knowledge in text augmentation.

\section{Limitations and future studies}

Although we are highly confident that observations made in this study are reliable, we were nevertheless unable to experiment with different initializations of the two REDA programs and different configurations of the classification models, constrained by available resources. Moreover, systematically and fairly evaluating a DA approach for NLP is uneasy or even unknown. The current study only illustrates a tip of the iceberg. 

In light of the limitations above, future studies may carry out similar experiments with differing setups, different NLP tasks, or even distinct methods of fusing a DA approach or technique with linguistic knowledge. Because of the simplicity and low cost of the five DA techniques employed in this study, it may also be important to re-examine the effectiveness and limitations of these random text editing operations for assorted NLP tasks. This may then give us some useful insights into building cheap and (highly) universal DA techniques for NLP, which is currently lacking in the field.

\bibliography{references}
\bibliographystyle{acl_natbib}

\end{document}

%% file: tables/lcqmc_data.tex
\begin{table}
\footnotesize
\centering
\begin{tabular}{cccc}
\hline
\textbf{Dataset} & \textbf{Total Pairs} & \textbf{Matched} & \textbf{Mismatched} \\ \hline
Train & 238,766 & 138,574 & 100,192 \\ \hline
Dev & 8,802 & 4,402 & 4,400 \\ \hline
Test & 12,500 & 6,250 & 6,250 \\ \hline
 \end{tabular}
\caption{\label{tab:data}
The basic statistics of LCQMC data sets.}
\end{table}

%% file: tables/quality_experiment.tex
\begin{table}
\footnotesize
\centering
\begin{tabular}{llccc}
\hline
&  & \textbf{One Edit} & \textbf{Two Edits} & \textbf{Three Edits} \\ \hline
SR & REDA & 22\% & 6\% & 2\% \\ 
& +N-gram & \textbf{88\%} & 79\% & 64\% \\ \hline
RS & REDA & 9\% & 4\% & 4\% \\ 
& +N-gram & \textbf{69\%} & 41\% & 34\% \\ \hline
RD & REDA & 16\% & 5\% & 2\% \\ 
& +N-gram & \textbf{39\%} & 22\% & 15\% \\ \hline
 \end{tabular}
\caption{\label{tab:quality_experiment}
The average accuracy scores of the two DA programs in three text restoration tasks based on different number of edits. SR: Synonym Replacement; RS: Random Swap; RD: Random Deletion.}
\end{table}

%% file: tables/aug_data.tex
\begin{table*}[h]
\footnotesize
\centering
\begin{tabular}{cccccc}
\hline
LCQMC & 5,000 &  10,000 & 50,000 & 100,000 & 238,766 \\ \hline
REDA & 66,267 & 132,513 & 563,228 & 929,176 & 2,218,512 \\ 
+N-gram & 64,358 & 128,649 & 544,583 & 893,779 & 2,133,163 \\ \hline
 \end{tabular}
\caption{\label{tab:aug_data}
The train set size for the corresponding REDA and $REDA_{+NG}$ augmented train sets.}
\end{table*}

%% file: tables/accu.tex
\begin{table*}[h]
\footnotesize
\centering
\begin{tabular}{lcccccc}
\hline
Models & 5k & 10k & 50k & 100k & Full Set & Average \\ \hline
BOW & 59.4\% & 60.4\% & 65.4\% & 67.8\% & 73.8\% & 65.4\% \\ 
+REDA & 58.1\% & 60.9\% & 68.2\% & 72.2\% & 76.4\% & 67.2\% \\ 
+$REDA_{+NG}$ & 58.8\% & 59.6\% & 68.1\% & 71.2\% & 76.0\% & 66.7\% \\ \hline

CNN & 59.3\% & 63.4\% & 67.2\% & 69.0\% & 72.9\% & 66.4\% \\ 
+REDA & 59.8\% & 62.6\% & 66.8\% & 69.8\% & 74.9\% & 66.8\% \\ 
+$REDA_{+NG}$ & 60.3\% & 62.0\% & 67.9\% & 69.1\% & 74.0\% & 66.7\% \\ \hline

LSTM & 60.0\% & 62.1\% & 66.2\% & 69.6\% & 74.8\% & 66.5\% \\ 
+REDA & 58.9\% & 61.5\% & 67.7\% & 71.8\% & 76.4\% & 67.3\% \\ 
+$REDA_{+NG}$ & 57.7\% & 60.9\% & 67.7\% & 71.7\% & 75.9\% & 66.8\% \\ \hline

GRU & 59.8\% & 61.9\% & 68.1\% & 70.3\% & 76.8\% & 67.4\% \\ 
+REDA & 58.7\% & 61.3\% & 68.7\% & 72.7\% & 76.8\% & 67.6\% \\ 
+$REDA_{+NG}$ & 58.8\% & 60.0\% & 67.8\% & 72.5\% & 76.6\% & 67.1\% \\ \hline

ERINE-Gram & \textbf{78.7\%} & \textbf{81.7\%} & \textbf{85.9\%} & \textbf{87.1\%} & \textbf{87.4\%} & \textbf{84.2\%} \\ 
+REDA & 77.5\% & 80.3\% & 84.1\% & 85.0\% & 85.7\% & 82.5\% \\ 
+$REDA_{+NG}$ & 78.6\% & 80.1\% & 83.8\% & 84.6\% & 85.8\% & 82.6\% \\ \hline

Average & 63.5\% & 65.9\% & 70.6\% & 72.8\% & 77.1\% & 70.0\% \\ 
+REDA & 62.6\% & 65.3\% & 71.1\% & 74.3\% & 78.0\% & 70.3\% \\ 
+$REDA_{+NG}$ & 62.8\% & 64.5\% & 71.1\% & 73.8\% & 77.7\% & 70.0\% \\ \hline
 \end{tabular}
\caption{\label{tab:accu}
Test set accuracy of the five classification models trained on the three types of train sets of varying size.}
\end{table*}

%% file: tables/metrics.tex
\begin{table*}[h]
\footnotesize
\centering
\begin{tabular}{l|ccc|ccc|ccc}
\hline

Models & \multicolumn{3}{c}{Baseline} \vline & \multicolumn{3}{c}{REDA} \vline & \multicolumn{3}{c}{$REDA_{+NG}$} \\ \hline

 &  Precision & Recall & F1  & Precision &  Recall & F1  & Precision & Recall & F1  \\ \hline

BOW &  61.5\% & 82.5\% & 70.4\%  & 63.3\% & 81.7\% & 71.3\% & 62.9\% & 81.8\% & 71.1\%  \\ \hline

CNN & 62.8\% & 80.5\% & 70.5\%  & 63.6\% &  78.1\% & 70.0\% & 63.8\% & 76.2\% & 69.3\%  \\ \hline

LSTM &  62.5\% & 82.7\% & 71.2\%  & 63.4\% & 81.4\% & 71.3\%  & 63.0\% & 82.1\% & 71.3\%  \\ \hline

GRU & 63.4\% & 82.4\% & 71.7\%  &  63.8\% & 81.9\% & 71.7\%  & 63.3\% & 81.7\% & 71.4\%  \\ \hline

ERINE-Gram &  \textbf{78.0\%} & 95.8\% & \textbf{85.9\%}  & 75.8\% & \textbf{95.9\%} &  84.6\%  & 76.0\% & 95.3\% & 84.6\%  \\ \hline

Average & 65.6\% &  84.8\% & 73.9\%  & 66.0\% & 83.8\% & 73.8\% & 65.8\% & 83.4\% & 73.5\%  \\ \hline

 \end{tabular}
\caption{\label{tab:metrics}
Average test set precision, recall, and F1 scores for the five classification models trained on the three types of train sets.}
\end{table*}

%% file: tables/qqqd_accu.tex
\begin{table*}[h]
\footnotesize
\centering
\begin{tabular}{lcccccc}
\hline
Models & 10k & 50k & 100k & 150k & Full Set (260k) & Average \\ \hline
BOW & 64.4\% & 69.9\% & 72.1\% & 74.2\% & 77.7\% & 71.7\% \\ 
+REDA & 62.5\% & 68.5\% & 71.6\% & 74.8\% & 78.0\% & 71.1\% \\ 
+$REDA_{+NG}$ & 62.9\% & 69.4\% & 74.0\% & 75.5\% & 78.2\% & 72.0\% \\ \hline

CNN & 66.1\% & 71.1\% & 72.6\% & 73.4\% & 75.9\% & 71.8\% \\ 
+REDA & 63.7\% & 69.9\% & 72.7\% & 75.3\% & 77.6\% & 71.8\% \\ 
+$REDA_{+NG}$ & 63.5\% & 69.3\% & 72.7\% & 74.7\% & 77.7\% & 71.6\% \\ \hline

LSTM & 65.7\% & \textbf{71.6}\% & 72.9\% & 75.0\% & 77.9\% & 72.6\% \\ 
+REDA & 64.0\% & 69.8\% & 72.5\% & 75.1\% & 78.1\% & 71.9\% \\ 
+$REDA_{+NG}$ & 64.9\% & 70.3\% & 72.7\% & 75.0\% & 78.1\% & 72.2\% \\ \hline

GRU & \textbf{67.2}\% & 71.0\% & \textbf{74.3}\% & 74.7\% & 77.4\% & 72.9\% \\ 
+REDA & 63.3\% & 70.0\% & 72.8\% & 74.8\% & 78.1\% & 71.8\% \\ 
+$REDA_{+NG}$ & 64.0\% & 70.2\% & 73.8\% & \textbf{75.7}\% & \textbf{78.9}\% & 72.5\% \\ \hline

Average & 65.9\% & 70.9\% & 73.0\% & 74.3\% & 77.2\% & 72.3\% \\ 
+REDA & 63.4\% & 69.6\% & 72.4\% & 75.0\% & 78.0\% & 71.7\% \\ 
+$REDA_{+NG}$ & 63.8\% & 69.8\% & 73.3\% & 75.2\% & 78.2\% & 72.1\% \\ \hline
 \end{tabular}
\caption{\label{tab:qqqd_accu}
Test set accuracy of four classification models trained on the three types of train sets of QQQD with varying sizes. Due to cost concerns, we did not finetune a pre-trained model, such as BERT, this time.}
\end{table*}

%% file: tables/qqqd_metrics.tex
\begin{table*}[h]
\footnotesize
\centering
\begin{tabular}{l|ccc|ccc|ccc}
\hline

Models & \multicolumn{3}{c}{Baseline} \vline & \multicolumn{3}{c}{REDA} \vline & \multicolumn{3}{c}{$REDA_{+NG}$} \\ \hline

 &  Precision & Recall & F1  & Precision &  Recall & F1  & Precision & Recall & F1  \\ \hline

BOW &  70.9\% & 73.5\% & 72.1\%  & 69.2\% & 76.1\% & 72.5\% & 71.1\% & 74.4\% & 72.7\%  \\ \hline

CNN & 70.5\% & 75.4\% & 72.8\%  & 70.7\% &  76.0\% & 73.1\% & 70.2\% & 76.5\% & 73.1\%  \\ \hline

LSTM &  70.5\% & \textbf{78.2}\% & \textbf{74.1}\%  & 70.5\% & 75.4\% & 72.8\%  & 71.4\% & 74.1\% & 72.7\%  \\ \hline

GRU & \textbf{71.8}\% & 75.5\% & 73.5\%  &  69.8\% & 76.9\% & 73.2\%  & 71.6\% & 74.5\% & 73.0\%  \\ \hline

Average & 70.9\% &  75.6\% & 73.1\%  & 70.1\% & 76.1\% & 72.9\% & 71.1\% & 74.9\% & 73.9\%  \\ \hline

 \end{tabular}
\caption{\label{tab:qqqd_metrics}
Average test set precision, recall, and F1 scores for the four classification models trained on the three types of train sets of QQQD.}
\end{table*}

%% file: acl_latex.bbl
\begin{thebibliography}{26}
\expandafter\ifx\csname natexlab\endcsname\relax\def\natexlab#1{#1}\fi

\bibitem[{Asai and Hajishirzi(2020)}]{asai-hajishirzi-2020-logic}
Akari Asai and Hannaneh Hajishirzi. 2020.
\newblock \href {https://doi.org/10.18653/v1/2020.acl-main.499} {Logic-guided
  data augmentation and regularization for consistent question answering}.
\newblock In \emph{Proceedings of the 58th Annual Meeting of the Association
  for Computational Linguistics}, pages 5642--5650, Online. Association for
  Computational Linguistics.

\bibitem[{Brants et~al.(2007)Brants, Popat, Xu, Och, and
  Dean}]{brants-etal-2007-large}
Thorsten Brants, Ashok~C. Popat, Peng Xu, Franz~J. Och, and Jeffrey Dean. 2007.
\newblock \href {https://aclanthology.org/D07-1090} {Large language models in
  machine translation}.
\newblock In \emph{Proceedings of the 2007 Joint Conference on Empirical
  Methods in Natural Language Processing and Computational Natural Language
  Learning ({EMNLP}-{C}o{NLL})}, pages 858--867, Prague, Czech Republic.
  Association for Computational Linguistics.

\bibitem[{Feng et~al.(2021)Feng, Gangal, Wei, Chandar, Vosoughi, Mitamura, and
  Hovy}]{feng-etal-2021-survey}
Steven~Y. Feng, Varun Gangal, Jason Wei, Sarath Chandar, Soroush Vosoughi,
  Teruko Mitamura, and Eduard Hovy. 2021.
\newblock \href {https://doi.org/10.18653/v1/2021.findings-acl.84} {A survey of
  data augmentation approaches for {NLP}}.
\newblock In \emph{Findings of the Association for Computational Linguistics:
  ACL-IJCNLP 2021}, pages 968--988, Online. Association for Computational
  Linguistics.

\bibitem[{Hou et~al.(2018)Hou, Liu, Che, and Liu}]{hou-etal-2018-sequence}
Yutai Hou, Yijia Liu, Wanxiang Che, and Ting Liu. 2018.
\newblock \href {https://aclanthology.org/C18-1105} {Sequence-to-sequence data
  augmentation for dialogue language understanding}.
\newblock In \emph{Proceedings of the 27th International Conference on
  Computational Linguistics}, pages 1234--1245, Santa Fe, New Mexico, USA.
  Association for Computational Linguistics.

\bibitem[{Iwana and Uchida(2021)}]{Iwana2021}
Brian~Kenji Iwana and Seiichi Uchida. 2021.
\newblock \href {https://doi.org/10.1371/journal.pone.0254841} {An empirical
  survey of data augmentation for time series classification with neural
  networks}.
\newblock \emph{PLOS ONE}, 16(7):1--32.

\bibitem[{Jurafsky and Martin(2009)}]{Jurafsky2009}
Daniel Jurafsky and James~H. Martin. 2009.
\newblock \emph{Speech and Language Processing: An Introduction to Natural
  Language Processing, Computational Linguistics, and Speech Recognition}, 2st
  edition.
\newblock Prentice Hall PTR, USA.

\bibitem[{Kang et~al.(2018)Kang, Khot, Sabharwal, and
  Hovy}]{kang-etal-2018-adventure}
Dongyeop Kang, Tushar Khot, Ashish Sabharwal, and Eduard Hovy. 2018.
\newblock \href {https://doi.org/10.18653/v1/P18-1225} {{A}dv{E}ntu{R}e:
  Adversarial training for textual entailment with knowledge-guided examples}.
\newblock In \emph{Proceedings of the 56th Annual Meeting of the Association
  for Computational Linguistics (Volume 1: Long Papers)}, pages 2418--2428,
  Melbourne, Australia. Association for Computational Linguistics.

\bibitem[{Kim(2014)}]{kim-2014-convolutional}
Yoon Kim. 2014.
\newblock \href {https://doi.org/10.3115/v1/D14-1181} {Convolutional neural
  networks for sentence classification}.
\newblock In \emph{Proceedings of the 2014 Conference on Empirical Methods in
  Natural Language Processing ({EMNLP})}, pages 1746--1751, Doha, Qatar.
  Association for Computational Linguistics.

\bibitem[{Kobayashi(2018)}]{kobayashi-2018-contextual}
Sosuke Kobayashi. 2018.
\newblock \href {https://doi.org/10.18653/v1/N18-2072} {Contextual
  augmentation: Data augmentation by words with paradigmatic relations}.
\newblock In \emph{Proceedings of the 2018 Conference of the North {A}merican
  Chapter of the Association for Computational Linguistics: Human Language
  Technologies, Volume 2 (Short Papers)}, pages 452--457, New Orleans,
  Louisiana. Association for Computational Linguistics.

\bibitem[{Kurata et~al.(2016)Kurata, Xiang, and Zhou}]{kurata16b_interspeech}
Gakuto Kurata, Bing Xiang, and Bowen Zhou. 2016.
\newblock \href {https://doi.org/10.21437/Interspeech.2016-727} {{Labeled Data
  Generation with Encoder-Decoder LSTM for Semantic Slot Filling}}.
\newblock In \emph{Proc. Interspeech 2016}, pages 725--729.

\bibitem[{Liu et~al.(2020)Liu, Wang, Xiang, and Meng}]{Liu:9240734}
Pei Liu, Xuemin Wang, Chao Xiang, and Weiye Meng. 2020.
\newblock \href {https://doi.org/10.1109/CCNS50731.2020.00049} {A survey of
  text data augmentation}.
\newblock In \emph{2020 International Conference on Computer Communication and
  Network Security (CCNS)}, pages 191--195.

\bibitem[{Liu et~al.(2018)Liu, Chen, Deng, Zeng, Chen, Li, and
  Tang}]{liu-etal-2018-lcqmc}
Xin Liu, Qingcai Chen, Chong Deng, Huajun Zeng, Jing Chen, Dongfang Li, and
  Buzhou Tang. 2018.
\newblock \href {https://aclanthology.org/C18-1166} {{LCQMC}:a large-scale
  {C}hinese question matching corpus}.
\newblock In \emph{Proceedings of the 27th International Conference on
  Computational Linguistics}, pages 1952--1962, Santa Fe, New Mexico, USA.
  Association for Computational Linguistics.

\bibitem[{Manning and Sch{\"u}tze(1999)}]{manning99foundations}
Christopher~D. Manning and Hinrich Sch{\"u}tze. 1999.
\newblock \href {http://nlp.stanford.edu/fsnlp/} {\emph{Foundations of
  Statistical Natural Language Processing}}.
\newblock The {MIT} Press, Cambridge, Massachusetts.

\bibitem[{Park et~al.(2019)Park, Chan, Zhang, Chiu, Zoph, Cubuk, and
  Le}]{Park2019}
Daniel~S. Park, William Chan, Yu~Zhang, Chung-Cheng Chiu, Barret Zoph, Ekin~D.
  Cubuk, and Quoc~V. Le. 2019.
\newblock \href {https://doi.org/10.21437/Interspeech.2019-2680} {{SpecAugment:
  A Simple Data Augmentation Method for Automatic Speech Recognition}}.
\newblock In \emph{Proc. Interspeech 2019}, pages 2613--2617.

\bibitem[{Sennrich et~al.(2016)Sennrich, Haddow, and
  Birch}]{sennrich-etal-2016-improving}
Rico Sennrich, Barry Haddow, and Alexandra Birch. 2016.
\newblock \href {https://doi.org/10.18653/v1/P16-1009} {Improving neural
  machine translation models with monolingual data}.
\newblock In \emph{Proceedings of the 54th Annual Meeting of the Association
  for Computational Linguistics (Volume 1: Long Papers)}, pages 86--96, Berlin,
  Germany. Association for Computational Linguistics.

\bibitem[{Severyn and Moschitti(2015)}]{Severyn2015}
Aliaksei Severyn and Alessandro Moschitti. 2015.
\newblock \href {https://doi.org/10.1145/2766462.2767738} {Learning to rank
  short text pairs with convolutional deep neural networks}.
\newblock In \emph{Proceedings of the 38th International ACM SIGIR Conference
  on Research and Development in Information Retrieval}, SIGIR '15, page
  373–382, New York, NY, USA. Association for Computing Machinery.

\bibitem[{Shorten and Khoshgoftaar(2019)}]{Shorten2019ASO}
Connor Shorten and Taghi~M. Khoshgoftaar. 2019.
\newblock A survey on image data augmentation for deep learning.
\newblock \emph{Journal of Big Data}, 6:1--48.

\bibitem[{Shorten et~al.(2021)Shorten, Khoshgoftaar, and Furht}]{Shorten2021}
Connor Shorten, Taghi~M. Khoshgoftaar, and Borko Furht. 2021.
\newblock Text data augmentation for deep learning.
\newblock \emph{Journal of Big Data}, 8:1--34.

\bibitem[{Singh et~al.(2019)Singh, McCann, Keskar, Xiong, and
  Socher}]{Singh2019}
Jasdeep Singh, Bryan McCann, Nitish~Shirish Keskar, Caiming Xiong, and Richard
  Socher. 2019.
\newblock \href {https://doi.org/10.48550/ARXIV.1905.11471} {Xlda:
  Cross-lingual data augmentation for natural language inference and question
  answering}.

\bibitem[{Tai et~al.(2015)Tai, Socher, and Manning}]{tai-etal-2015-improved}
Kai~Sheng Tai, Richard Socher, and Christopher~D. Manning. 2015.
\newblock \href {https://doi.org/10.3115/v1/P15-1150} {Improved semantic
  representations from tree-structured long short-term memory networks}.
\newblock In \emph{Proceedings of the 53rd Annual Meeting of the Association
  for Computational Linguistics and the 7th International Joint Conference on
  Natural Language Processing (Volume 1: Long Papers)}, pages 1556--1566,
  Beijing, China. Association for Computational Linguistics.

\bibitem[{Tien et~al.(2019)Tien, Le, Tomohiro, and Tatsuya}]{TIEN2019102090}
Nguyen~Huy Tien, Nguyen~Minh Le, Yamasaki Tomohiro, and Izuha Tatsuya. 2019.
\newblock \href {https://doi.org/https://doi.org/10.1016/j.ipm.2019.102090}
  {Sentence modeling via multiple word embeddings and multi-level comparison
  for semantic textual similarity}.
\newblock \emph{Information Processing \& Management}, 56(6):102090.

\bibitem[{Wang and Yang(2015)}]{wang-yang-2015-thats}
William~Yang Wang and Diyi Yang. 2015.
\newblock \href {https://doi.org/10.18653/v1/D15-1306} {That{'}s so
  annoying!!!: A lexical and frame-semantic embedding based data augmentation
  approach to automatic categorization of annoying behaviors using {\#}petpeeve
  tweets}.
\newblock In \emph{Proceedings of the 2015 Conference on Empirical Methods in
  Natural Language Processing}, pages 2557--2563, Lisbon, Portugal. Association
  for Computational Linguistics.

\bibitem[{Wei and Zou(2019)}]{wei-zou-2019-eda}
Jason Wei and Kai Zou. 2019.
\newblock \href {https://doi.org/10.18653/v1/D19-1670} {{EDA}: Easy data
  augmentation techniques for boosting performance on text classification
  tasks}.
\newblock In \emph{Proceedings of the 2019 Conference on Empirical Methods in
  Natural Language Processing and the 9th International Joint Conference on
  Natural Language Processing (EMNLP-IJCNLP)}, pages 6382--6388, Hong Kong,
  China. Association for Computational Linguistics.

\bibitem[{Xiao et~al.(2020)Xiao, Li, Zhang, Sun, Tian, Wu, and
  Wang}]{xiao2020ernie-gram}
Dongling Xiao, Yu-Kun Li, Han Zhang, Yu~Sun, Hao Tian, Hua Wu, and Haifeng
  Wang. 2020.
\newblock Ernie-gram: Pre-training with explicitly n-gram masked language
  modeling for natural language understanding.
\newblock arXiv.

\bibitem[{Xie et~al.(2020)Xie, Dai, Hovy, Luong, and Le}]{xie2020unsupervised}
Qizhe Xie, Zihang Dai, Eduard Hovy, Thang Luong, and Quoc Le. 2020.
\newblock \href
  {https://proceedings.neurips.cc/paper/2020/file/44feb0096faa8326192570788b38c1d1-Paper.pdf}
  {Unsupervised data augmentation for consistency training}.
\newblock In \emph{Advances in Neural Information Processing Systems},
  volume~33, pages 6256--6268. Curran Associates, Inc.

\bibitem[{Zhang et~al.(2015)Zhang, Zhao, and LeCun}]{Zhang2015}
Xiang Zhang, Junbo Zhao, and Yann LeCun. 2015.
\newblock \href
  {https://proceedings.neurips.cc/paper/2015/file/250cf8b51c773f3f8dc8b4be867a9a02-Paper.pdf}
  {Character-level convolutional networks for text classification}.
\newblock In \emph{Advances in Neural Information Processing Systems},
  volume~28. Curran Associates, Inc.

\end{thebibliography}
